\documentclass{article}

\usepackage[preprint]{neurips_2026}

\usepackage[utf8]{inputenc} 
\usepackage[T1]{fontenc}    
\usepackage{hyperref}       
\usepackage{url}            
\usepackage{booktabs}       
\usepackage{amsfonts}       
\usepackage{nicefrac}       
\usepackage{microtype}      
\usepackage{xcolor}         
\usepackage{graphicx} 
\usepackage{float}
\usepackage{amsmath}
\usepackage{booktabs}
\usepackage{array}

\restylefloat{table}
\restylefloat{figure}

\title{Mental Health AI Safety Claims Must Preserve Temporal Evidence}

\author{
  \And
  Srimonti Dutta\\
  Indian Institute of Technology Guwahati\\
  \texttt{\small srimonti.dutta@gmail.com} \\
  \And
  Ratna Kandala \\
  University of Kansas\\
  \texttt{\small ratnanirupama@gmail.com} \\
  }

\begin{document}

\maketitle

\begin{abstract}
The safety of mental health AI is often judged at the wrong temporal scale. Current evaluations typically score isolated responses, endpoint outcomes, or aggregate dialogue quality, while clinically consequential failures may arise from the order and accumulation of interactions themselves, including delayed escalation, repeated reinforcement, dependency formation, failed repair, and gradual deterioration across turns. This paper argues that this mismatch is not merely a limitation of evaluation coverage but a source of invalid safety conclusions. We introduce \textit{Temporal Safety Non-Identifiability}, a formal account of why safety properties that depend on sequence, timing, accumulation, or recovery cannot be certified by protocols that discard those features. From this formalization, we develop SCOPE (Safety Claims Over Preserved Evidence) as a general principle for aligning safety claims with the evidence an evaluation actually retains, and instantiate it as SCOPE-MH, a mental-health instantiation of this reporting standard. We operationalize SCOPE-MH through a proof-of-concept on the Anno-MI dataset of expert-annotated motivational interviewing conversations, which reveals mechanisms of failure that per-turn behavior scoring does not represent. We propose SCOPE-MH as a diagnostic complement to existing evaluation infrastructure and argue that evaluation preserving temporal evidence is necessary, not optional, for safety-critical mental health AI deployment.
\end{abstract}

\section{Introduction}

The limitations of endpoint measurement are well recognized in mental health research~\citep{moreno2026umbrella}. Cross-sectional snapshots and final outcomes can be clinically useful, but they often miss patterns of change, deterioration, relapse, recovery, and accumulating risk that are visible only over time. Mental health AI has not yet fully absorbed this lesson. \citet{morrin2026journey} argue that risk in mental health chatbots does not arise only at a final tipping point, but through temporal effects that accumulate across extended dialogue. We sharpen this argument into a claim about evaluation validity: when a safety property depends on temporal structure, an evaluation that discards that structure cannot certify the property it claims to measure.

\textbf{Our position is that safety claims for mental health AI are valid only when the evaluation preserves the temporal evidence required to assess the claimed safety property.} Endpoint outcomes, isolated response judgments, and aggregate dialogue scores remain useful, but they license only bounded claims. They can show that a user improved by follow-up, that a response looked appropriate in isolation, or that average dialogue quality was high. They cannot, by themselves, show that a system intervened at the right time, avoided cumulative reinforcement of harm, maintained a coherent stance as risk evolved, recovered after a miscalibrated response, or referred the user appropriately. The central question for interactive AI evaluation is therefore not only whether a system passed a benchmark, but what safety claims the benchmark's evidence is capable of supporting.

The scale of deployment makes this problem urgent. \citet{mcbain2025use} find that nearly one in eight US adolescents and young adults now use AI chatbots for mental health advice, with one in five among those aged 18 to 21. \citet{rousmaniere2025large} report that 48.7\% of adults with ongoing mental health conditions use major LLMs therapeutically. \citet{hua2025charting} document that LLM-based architectures surged to 45\% of mental health AI studies in 2024, yet only 16\% underwent clinical efficacy testing. This direction is consistent with broader mental health research emphasizing personalized temporal monitoring and dynamic recovery rather than isolated symptom states~\citep{stringer2026,bosl2025dynamical,ngabo2025prognostic}. A system that acts over time cannot be adequately assessed only by methods that observe final outcomes, isolated responses, or aggregate scores.

Mental health AI inherits evaluation practices from two mature traditions, neither designed to validate temporally extended interactive systems. Clinical evaluation compresses interaction histories into endpoint outcomes such as symptom reduction, adherence, satisfaction, and adverse events; the Woebot randomized controlled trial, for example, evaluated a conversational Cognitive Behavioral Therapy (CBT) intervention using PHQ-9, GAD-7, and PANAS at baseline and follow-up~\citep{fitzpatrick2017delivering}, and systematic reviews similarly synthesize evidence through depression, anxiety, distress, and well-being outcomes~\citep{li2023systematic}. Machine learning evaluation introduces a different compression: fixed prompts, independent test instances, rubric judgments, preference scores, and aggregate benchmark performance, as in MMLU and HELM~\citep{hendrycks2020measuring,liang2023holisticevaluationlanguagemodels}. These methods are valuable, but mental health support rarely has the structure of an independent input-output task. The meaning of a user's message may depend on earlier hesitation, repeated reassurance-seeking, gradual disclosure, or escalating risk; the appropriateness of a response depends on where it occurs in the interaction.

Recent work has begun to address this gap through safety-oriented benchmark questions, LLM-based evaluation tools, realistic counseling scenarios, multi-turn mental health support benchmarks, risk-specific dashboards, role-aware adversarial interaction evaluation, and memory-based detection of manipulative patterns~\citep{park2025buildingtrustmentalhealth,li2025counselbench,pombal2025mindeval,zhang2026mhdash,lee2026mhsafeeval,kandala2026echoguard}. These are important steps, but multi-turn coverage alone does not solve the validity problem. A benchmark may contain dialogue and still fail to support the intended safety claim if it reports only final responses, average scores, or event counts while discarding the temporal evidence on which the claim depends. The problem is not merely benchmark coverage. It is claim validity.

We formalize this interaction mismatch as \textit{Temporal Safety Non-Identifiability}. If an endpoint-style protocol observes an interaction only through a compression function \(\phi:\mathcal{T}\rightarrow\mathcal{Z}\), then safety becomes non-identifiable whenever two temporally distinct interactions with different safety statuses produce the same compressed representation. No downstream scoring rule can recover a distinction that the evaluation protocol has erased. From this formalization, we introduce \textit{SCOPE: Safety Claims Over Preserved Evidence}, a reporting and diagnostic standard for aligning safety claims with the evidence an evaluation actually retains. We instantiate this standard as \textit{SCOPE-MH} for mental health AI, where temporal safety commonly depends on longitudinal consistency, harm accumulation, intervention timing, recovery capability, and referral correctness.

This paper makes four contributions. First, we identify the claim-evidence mismatch in mental health AI evaluation: the use of endpoint outcomes, per-turn scores, or aggregate dialogue ratings to support safety claims whose determining evidence is temporal. Second, we formalize this mismatch as Temporal Safety Non-Identifiability. Third, we introduce SCOPE as a reusable reporting standard and instantiate it as SCOPE-MH for mental health. Fourth, we provide a proof-of-concept audit on Anno-MI~\citep{wu2022anno}, analyzing 131 expert-annotated motivational interviewing conversations. The empirical study does not aim to show that one simple temporal metric outperforms expert-coded per-turn scoring. It shows that temporal evidence changes what an evaluation can observe, explain, and validly claim.

The position defended here is not that endpoint outcomes or response-level benchmarks should be abandoned. The position is that their claims must be bounded by the evidence they preserve. Endpoint outcomes can support claims about terminal user change. Response-level benchmarks can support claims about local output quality. Aggregate dialogue scores can support limited claims about average behavior. But claims about delayed escalation, cumulative reinforcement, dependency formation, failed repair, or referral correctness require temporal evidence. When mental health AI safety claims depend on time, evaluation must preserve time.
\vspace{- 3mm}

\section{The Claim-Evidence Mismatch in Interactive AI Evaluation}
\label{sec:claim-evidence-mismatch}

Safety evaluation is not only a question of measurement accuracy. It is also a question of claim validity. An evaluation may measure something reliably and still fail to support the safety claim being made. This occurs when the evidence collected by the evaluation is too compressed, too local, or too temporally coarse to identify the property being asserted.

We call this the \textit{claim-evidence mismatch}: the use of endpoint outcomes, isolated response judgments, aggregate transcript scores, event counts, or averaged multi-turn ratings to support safety claims whose determining evidence lies in temporal structure. Endpoint-style evidence is not useless. Clinical endpoint outcomes are indispensable for evaluating whether users improve or deteriorate during and after use. Response-level judgments are necessary because a system that gives unsafe advice or fails to recognize crisis language is unsafe in an obvious way. Aggregate dialogue scores and event counts may also capture useful facts. The problem begins when these forms of evidence are used to support claims beyond their evidential scope. Endpoint outcomes can support claims about terminal user change, but not about how that outcome was produced. Isolated response scores can support claims about local response quality, but not about whether the system remains safe as user state evolves. Event counts can show whether something occurred, but not whether it occurred at the right severity, time, or manner.

A safety property is temporal when its truth depends on the ordered path of interaction rather than only on terminal or aggregate quantities. In mental health AI, safety may depend on whether the system intervened at the first credible risk cue, whether repeated reassurance increased avoidance or dependency, whether the system maintained a coherent stance as severity changed, whether it recovered after a miscalibrated response, or whether referral occurred at the right severity and time. The same response can therefore have different safety meanings depending on when it occurs. Reflective validation may be helpful when a user first expresses sadness, but unsafe if repeated after escalating hopelessness and disclosure of intent. A crisis resource may be appropriate in isolation, but inadequate if it appears only after several earlier escalation cues were missed.

Formally, let an interaction trajectory be an ordered sequence of user states, system actions, contextual information, and risk-relevant variables over an interaction horizon \(H\):
\[
\tau_{1:H}=\{(x_t,a_t,h_t,r_t)\}_{t=1}^{H}.
\]
Here, \(x_t\) denotes the user's observable input or expressed state, \(a_t\) the system action, \(h_t\) the available history and context, and \(r_t\) a risk-relevant state, whether observed or inferred. Such states may include distress, suicidal ideation, dependency, disengagement, impaired reality testing, crisis severity, or need for referral. We use \textit{trajectory} for the mathematical object; the safety properties of interest are temporal because their truth depends on ordering, timing, accumulation, adaptation, and recovery within that trajectory.

\subsection{Temporal Safety Non-Identifiability}

The claim-evidence mismatch can be stated as a non-identifiability problem. Let \(\mathcal{T}\) denote the set of possible interaction trajectories, and let \(S(\tau)\) denote the safety status of \(\tau\) with respect to a specified safety predicate. In practice, \(S\) may be binary, graded, probabilistic, multidimensional, or expert-adjudicated. In probabilistic settings, \(S(\tau)\) can be read as the risk distribution, expected risk, or calibrated probability assigned to \(\tau\).

An endpoint-style evaluation protocol observes the interaction only through a compression function
\[
\phi:\mathcal{T}\rightarrow\mathcal{Z},
\]
where \(\phi(\tau)\) may include a final symptom score, average response rating, satisfaction score, engagement summary, adverse-event count, final answer, aggregate transcript score, event count, or benchmark performance measure. Any evaluator relying only on this representation can be written as \(E_{\mathrm{end}}(\tau)=g(\phi(\tau))\).

\paragraph{Definition 1: Temporal Safety Non-Identifiability.}
An endpoint-style protocol exhibits \textit{Temporal Safety Non-Identifiability} for safety predicate \(S\) when there exist two trajectories \(\tau,\tau'\in\mathcal{T}\) such that
\[
\phi(\tau)=\phi(\tau') \quad \text{but} \quad S(\tau)\neq S(\tau').
\]

\paragraph{Proposition 1.}
If such trajectories exist, no deterministic evaluator that depends only on \(\phi(\tau)\) can correctly determine \(S\) for all trajectories in \(\mathcal{T}\). Likewise, no randomized evaluator whose output distribution depends only on \(\phi(\tau)\) can guarantee correct determination for both trajectories.

\paragraph{Proof sketch.}
Since \(\phi(\tau)=\phi(\tau')\), every evaluator that receives only \(\phi\) receives identical input for both trajectories. It must therefore assign the same score, label, or decision to both. If \(S(\tau)\neq S(\tau')\), at least one assignment must be wrong. The failure is not due to insufficient sample size, weak calibration, poor annotators, or an inadequate scoring rule. It follows from information loss. 

The implication is simple but consequential: a safety property is identifiable from an evaluation representation only if the property is constant over the equivalence classes induced by that representation. More data of the same compressed form do not solve the problem. Nor does replacing a human evaluator with a stronger model. The missing distinction is not present in the evidence.

\subsection{How the Mismatch Produces False Assurance}

Temporal Safety Non-Identifiability appears in practice through three recurring failure modes. \textit{Unit-of-analysis mismatch} treats acceptable isolated responses as evidence of safe interactive behavior. \textit{Temporal evidence loss} removes the timing, ordering, escalation latency, accumulation, adaptation, and recovery patterns through which many risks emerge. The \textit{outcome attribution gap} treats favorable endpoints such as symptom reduction, satisfaction, engagement, or adverse-event counts as evidence of safe system conduct without showing that the system caused or safely supported those outcomes. These mechanisms differ, but they produce the same error: safety claims that exceed the evidence. SCOPE makes these evidential boundaries explicit; Appendix~\ref{app:false-assurance} gives a fuller diagnostic table.

\section{SCOPE: Safety Claims Over Preserved Evidence}
\label{sec:scope}

The previous section showed why temporally compressed evidence cannot certify temporal safety properties. The remedy is not simply longer benchmarks, more annotators, or more aggregate metrics. The more basic requirement is claim discipline: an evaluation should only support safety claims for properties whose determining evidence it preserves.

We call this principle \textsc{SCOPE}: \textit{Safety Claims Over Preserved Evidence}. SCOPE is a reporting and diagnostic standard for interactive AI safety evaluation. It does not prescribe a single metric, horizon, dataset, or model judge. Instead, it requires evaluators to make explicit the relationship between the safety claim being asserted and the evidence the evaluation actually retains. A one-turn refusal benchmark may support a claim about local refusal behavior. A post-use symptom measure may support a claim about terminal user change. An aggregate transcript score may support a limited claim about average interaction quality. None of these, by itself, establishes that the system detected risk at the right time, avoided cumulative harm, recovered after miscalibration, or referred appropriately.

\begin{table}[t]
\centering
\caption{The \textsc{SCOPE} Reporting Standard. Interactive AI safety evaluations should report the claim being made, the temporal evidence required to assess it, the evidence preserved, and the claims that remain unsupported.}
\label{tab:scope-standard}
\small
\setlength{\tabcolsep}{4pt}
\renewcommand{\arraystretch}{1.08}
\begin{tabular}{@{}p{0.27\linewidth}p{0.65\linewidth}@{}}
\toprule
\textbf{SCOPE field} & \textbf{Required disclosure} \\
\midrule
Safety claim & What safety property is being asserted? \\
Evaluation horizon & Was safety assessed over one turn, one dialogue, multiple sessions, or longitudinal use? \\
Unit of analysis & Is the evaluation judging a response, dialogue, user, episode, system version, or deployment population? \\
Temporal determinants & Does the claim depend on timing, ordering, accumulation, adaptation, recovery, referral, or user-state change? \\
Preserved evidence & Which temporal determinants are actually observed by the evaluation protocol? \\
Unsupported claims & What stronger claims does the evaluation not license? \\
Outcome linkage & Are endpoint outcomes linked to interactional evidence, or only measured before and after use? \\
Privacy constraints & What temporal interaction data are collected, retained, anonymized, aggregated, or protected? \\
\bottomrule
\end{tabular}
\end{table}

SCOPE is not a replacement for existing evaluation tools; it is a constraint on how their results may be interpreted. If an evaluation observes only isolated responses, it may support claims about local response quality, but not longitudinal consistency. If it measures only final symptom change, it may support claims about terminal user state, but not whether the system behaved safely during the interaction. If it counts whether referral occurred but not when, why, or how, it may support a claim that referral was mentioned, but not that referral was correct. Table~\ref{tab:scope-standard} gives the minimal reporting fields.

\subsection{SCOPE-MH: Mental Health AI Instantiation}

We instantiate SCOPE for mental health AI as \textsc{SCOPE-MH}. Mental health AI is a natural stress test for the standard because many safety properties are temporal: risk can emerge through delay, repetition, failed repair, inappropriate reassurance, dependency, or mistimed referral. SCOPE-MH identifies five diagnostic dimensions that endpoint-style evaluation commonly fails to certify. These dimensions are not exhaustive, but they provide a minimal vocabulary for asking whether an evaluation preserves the temporal evidence needed to support common safety claims. Table~\ref{tab:scope-mh-dimensions} summarizes these dimensions and minimal operationalizations.

\begin{table}[t]
\centering
\caption{Diagnostic dimensions in \textsc{SCOPE-MH}. Each dimension identifies a temporal safety property that endpoint-style evaluation can fail to certify.}
\label{tab:scope-mh-dimensions}
\small
\setlength{\tabcolsep}{3.5pt}
\renewcommand{\arraystretch}{1.08}
\begin{tabular}{@{}p{0.23\linewidth}p{0.35\linewidth}p{0.34\linewidth}@{}}
\toprule
\textbf{Dimension} & \textbf{What endpoint-style evaluation misses} & \textbf{Minimal operationalization} \\
\midrule
Longitudinal consistency & Whether the system maintains a coherent and safe stance as context evolves. & Track stance shifts, contradictions, boundary drift, and policy inconsistency. \\
Harm accumulation & Whether locally acceptable responses compound into risk. & Track repeated reinforcement, dependency signals, cumulative risk markers, or worsening user-state indicators. \\
Intervention timing & Whether the system intervenes when risk first becomes detectable. & Measure escalation latency from first risk cue to appropriate intervention. \\
Recovery capability & Whether the system repairs after unsafe, delayed, or miscalibrated responses. & Track acknowledgment, correction, redirection, stabilization, and subsequent user-state change. \\
Referral correctness & Whether referral occurs at the right severity, time, and manner. & Assess referral timing, specificity, severity alignment, continuity, and unsafe disengagement. \\
\bottomrule
\end{tabular}
\end{table}

SCOPE-MH is not a certificate of global safety. It does not imply that every system must be evaluated over unlimited horizons, replace clinical outcomes with process metrics, or assume that transcripts alone reveal user welfare. The claim is narrower: when a safety property depends on temporal structure, the evaluation must preserve enough of that structure to assess it. A crisis-support system, a long-term companion system, and a therapy-assistive system should not be evaluated with identical horizons or identical safety predicates. What remains constant is the reporting discipline: do not claim safety for a property whose determining evidence the protocol does not observe.

\section{SCOPE-MH in Practice: A Proof-of-Concept on Expert-Annotated Dialogues}
\label{sec:empirical}

To ground the argument empirically, we operationalize SCOPE-MH on Anno-MI~\citep{wu2022anno}, a dataset of 133 motivational interviewing conversations annotated by trained MI coders. \footnote{In deployment, such labels would need to be produced by trained coders or automated classifiers; here we use existing Anno-MI annotations to isolate the evaluation question.} Each utterance carries speaker identity (therapist or client), therapist behavior \(\in\) \{\texttt{reflection}, \texttt{question}, \texttt{therapist\_input}, \texttt{other}\}, and \texttt{client\_talk\_type} \(\in\) \{\texttt{change}, \texttt{neutral}, \texttt{sustain}\}. Each conversation is labeled \textit{high} or \textit{low} quality by expert coders, providing an expert reference label against which evidence sources can be compared. We chose Anno-MI for three reasons: it contains real multi-turn conversations, conversation-level expert quality labels, and turn-level therapist behavior and client talk-type annotations fine-grained enough to operationalize both per-turn and temporal evidence sources directly from the data.

After excluding two conversations with fewer than four labeled client utterances, our analysis covers 131 conversations: 110 high-quality and 21 low-quality. For each conversation, utterances are grouped and sorted by temporal order, then separated into therapist and client turns.\footnote{\url{https://github.com/ratnakandala/mental-health-trajectories}} The \texttt{client\_talk\_type} labels directly encode motivational direction, making temporal client-language patterns computable from the annotation without additional modeling.

The goal is not to establish a deployment-ready detector or to show that one simple temporal metric outperforms expert-coded per-turn scoring. The goal is to audit what different evidence sources can observe, what claims they can support, and where their blind spots remain. We compare a \textit{per-turn behavior baseline}, defined as the fraction of therapist turns labeled reflection or question, with temporal client-language signals derived from change-talk and sustain-talk over the conversation. In deployment, both therapist behavior labels and client talk-type labels would need to be produced by trained coders or automated classifiers; here we use existing Anno-MI annotations to isolate the evaluation-validity question.

\subsection{RQ1: Do Temporal Signals Contain Quality-Relevant Information?}

The per-turn behavior baseline is the strongest classifier in this dataset, separating high- and low-quality conversations with \(d=1.599\) and \(p<0.001\). This is expected: Anno-MI provides expert-coded therapist behavior labels that are closely tied to motivational interviewing quality. Sustain-talk delta is the strongest temporal signal. It separates high- and low-quality conversations under a parametric test (\(p=0.037,d=0.462\)), although robustness analyses in the Appendix suggest treating this result as preliminary. The sign is interpretable: high-quality conversations show decreasing resistance across the interaction, while low-quality conversations show increasing resistance. This directional client-language evidence is not represented by the per-turn behavior score. Table~\ref{tab:empirical-summary} summarizes the main empirical results.

\subsection{RQ2: What Operating Profile Do Temporal Signals Have?}

Threshold detection should be interpreted as illustrative, not deployment-validated. The full-conversation temporal threshold \(\theta=-0.10\) is selected by F1 maximization across thresholds from \(-0.05\) to \(-0.50\). The early-warning threshold is separately selected on the early-warning task; because both are selected on the same dataset used for evaluation, threshold-free ROC, precision-recall, bootstrap, and sensitivity analyses are reported in the Appendix.

\begin{table}[t]
\centering
\caption{Compact summary of the Anno-MI proof-of-concept. Thresholded operating points are illustrative because thresholds are selected on the same dataset.}
\label{tab:empirical-summary}
\small
\setlength{\tabcolsep}{3.2pt}
\renewcommand{\arraystretch}{1.08}
\begin{tabular}{@{}p{0.27\linewidth}p{0.31\linewidth}p{0.34\linewidth}@{}}
\toprule
\textbf{Question} & \textbf{Result} & \textbf{Interpretation} \\
\midrule
Group separation & Per-turn baseline: \(d=1.599,p<0.001\). Sustain-talk delta: \(d=0.462,p=0.037\). & Per-turn scoring is the stronger classifier; temporal client-language change carries quality-relevant but preliminary signal. \\
Full-conversation detection & Per-turn baseline: 19/21 detected, 55/110 false alarms. Temporal signal: 7/21 detected, 13/110 false alarms. Combined: 19/21 detected. & The temporal signal does not add recall, but behaves as a more selective audit trigger at this operating point. \\
Midpoint warning & Per-turn baseline: 13/21 detected. Temporal signal: 9/21 detected. Combined: 14/21 detected, with higher false alarms. & Temporal evidence provides limited complementary early-warning coverage, not a validated intervention policy. \\
\bottomrule
\end{tabular}
\end{table}

The empirical value of the temporal signal is therefore not superiority as a detector. Its value is evidential: it preserves a different object, the direction of client motivational language over time. In full-conversation detection, it is lower recall but more selective. At midpoint, it adds one low-quality conversation missed by the per-turn baseline, but at a false-alarm cost. These results support SCOPE-MH as a claim-validity framework rather than a single-metric benchmark.

\subsection{RQ3: What Do Temporal Signals Reveal and Miss?}

\begin{figure}[htbp]
  \centering
  \includegraphics[width=\linewidth]{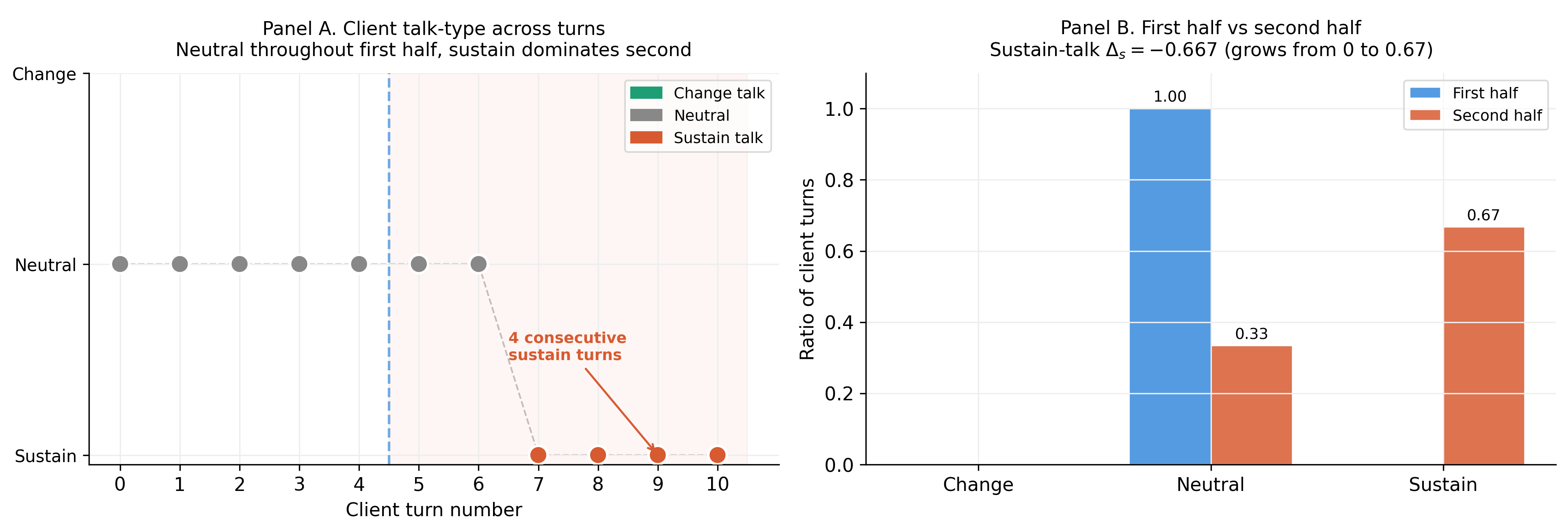}
  \caption{
    \textbf{Conversation 27 temporal trace reveals the mechanism of failure.}
    The per-turn behavior baseline and the temporal signal both flag the conversation as low quality, but they preserve different evidence. The per-turn score reflects sparse use of reflection or questioning. The temporal trace shows the mechanism: the first half is neutral, while four consecutive sustain-talk turns emerge in the second half after a directive intervention. This illustrates the SCOPE claim that a correct verdict does not by itself preserve the evidence needed to explain the safety-relevant temporal pattern.
  }
  \label{fig:conv27}
\end{figure}

Case analysis shows why verdict accuracy is not the same as evidential adequacy. In Conversation 27, an asthma-management session, both the per-turn behavior baseline (score \(=0.083\)) and the temporal signal (\(\Delta s=-0.667\)) correctly flag the conversation as low quality. But they support different claims. The per-turn score identifies poor local MI practice because the therapist rarely uses reflection or questioning. The temporal trace shows the mechanism: the first half consists entirely of neutral client responses; after the therapist's directive to remove the client's dog of twelve years, four consecutive sustain-talk turns emerge. Sustain-talk rises from 0.00 to 0.67 while change-talk remains absent. The correct verdict alone does not show this temporal mechanism. Figure~\ref{fig:conv27} shows the temporal trace that does. Table~\ref{tab:scope-error-analysis} summarizes the SCOPE-MH interpretation of these cases.

\begin{table}[t]
\centering
\caption{SCOPE-MH error analysis of three low-quality Anno-MI conversations. Temporal evidence can reveal mechanisms, but simple temporal metrics do not capture all temporal failure modes.}
\label{tab:scope-error-analysis}
\small
\setlength{\tabcolsep}{3.2pt}
\renewcommand{\arraystretch}{1.08}
\begin{tabular}{@{}p{0.09\linewidth}p{0.16\linewidth}p{0.16\linewidth}p{0.51\linewidth}@{}}
\toprule
\textbf{Case} & \textbf{Per-turn baseline} & \textbf{Temporal signal} & \textbf{SCOPE implication} \\
\midrule
27 & Fail & Fail & Correct verdict does not imply adequate evidence: the temporal trace reveals resistance accumulation after a directive intervention. \\
53 & Pass & Pass & Simple sustain-talk delta misses stagnation: a long neutral plateau can indicate lack of therapeutic progress without overt resistance growth. \\
131 & Pass & Pass & Simple sustain-talk delta misses backsliding: change-talk can deteriorate even when sustain-talk does not increase. \\
\bottomrule
\end{tabular}
\end{table}

These cases show why SCOPE-MH is not a single-metric proposal. Sustain-talk delta preserves evidence about resistance accumulation, but not all temporal failure modes. Stagnation and backsliding require different temporal signals. This is precisely the role of SCOPE: to state what a metric preserves, what it can support, and what remains unidentified.

\section{Discussion}
\label{sec:discussion}

\subsection{Alternative Views and Objections}

\textbf{Endpoint-Style Evaluation Is Sufficient in Practice} The most direct objection is empirical: on Anno-MI, per-turn process score achieves strong separation between quality groups ($d = 1.599$, $p < 0.001$), outperforming temporal metrics as a standalone predictors. We offer three responses. First, Anno-MI's per-turn behavior features are unusually rich, each turn is annotated by trained MI coders, a resource unlikely to be available at scale in routine AI evaluation pipelines. 
Both evidence sources are annotation-dependent in this proof-of-concept; the contrast is not annotation-free versus annotation-heavy evaluation, but the type of evidence preserved. In practice, routine AI evaluation often relies on automated scoring or LLM judges that capture far less signal. Second, the cases both methods miss show that some failures are not captured by either local therapist behavior or simple sustain-talk delta; stagnation and backsliding require different temporal operationalizations. Third, evaluation failures in mental health AI are asymmetric: at the selected illustrative full-conversation operating point, the temporal signal has a false alarm rate of 12\% versus 50\% for the per-turn behavior baseline, and combined evaluation at midpoint detects 14 of 21 low-quality conversations (67\%) versus 13/21 (62\%) for the per-turn baseline.

\textbf{Temporal Evidence Is Too Complex.}
Given client talk-type labels, sustain-talk delta requires only a single arithmetic operation: no model training, no LLM judge, and no additional annotation beyond those labels. The substantive difficulty is obtaining reliable talk-type labels at scale, but talk-type classification is well studied and lightweight classifiers trained on MI corpora achieve high accuracy~\citep{tanana2016comparison}.

\textbf{This Is Just Sequential Evaluation Reframed} The distinction between prior work and SCOPE-MH is diagnostic intent. MindEval~\citep{pombal2025mindeval} and MHDash~\citep{zhang2026mhdash}
ask what happened. SCOPE asks what claims become invalid when the evaluation discards temporal evidence. This produces different evaluation targets, different
metrics, and different failure modes. The proof-of-concept focuses on client-side temporal signals and arc-level deterioration, but SCOPE-MH itself is broader: it also includes system-side timing, recovery, referral, and longitudinal consistency.

\section{Limitations}
\label{sec:limitations}
Anno-MI contains only 21 low-quality conversations, limiting
statistical power; timing and persistence metrics trend in the
expected direction, but do not reach significance after correction, which we interpret as a power limitation rather than a null effect. The dataset covers motivational interviewing only. Sustain-talk delta is meaningful here because talk-type labels directly encode motivational direction, but analogous signals in crisis support,
CBT, or companionship contexts require separate operationalization. SCOPE-MH is a diagnostic framework, not a deployment-ready protocol; translating it into precise thresholds and validation standards requires additional empirical and normative work, including clinical validation. Our operationalization depends on the MITI-derived talk-type
labels~\citep{moyers2016motivational}; fully automated pipelines operating on raw logs remain an open problem. The thresholds $\theta = -0.10$ and $\theta_{ew} = -0.15$ were selected by F1 maximisation on the same dataset used for evaluation and should be treated as illustrative; sensitivity analysis is in Appendix. Our results establish an association between temporal signals and expert quality labels, not causation; prospective studies with outcome measures are needed~\citep{morrin2026journey}. Finally, all experiments use human therapist data; whether the failure modes we identify match those of deployed AI systems is an open empirical question, though the theoretical argument holds regardless: if safety depends on temporal structure, endpoint-style evidence cannot certify it.

\section{Ethical Considerations}
\label{sec:ethics}

Temporal-evidence evaluation, if poorly calibrated, can produce false alarms or miss failures in new interaction types; deployers should treat temporal signals as one input among several, not as a binary safety gate. Deployers who use only endpoint-style evidence should not make temporal safety claims that their evidence does not support. We urge explicit acknowledgment of the evaluation horizon as a first-class reporting requirement alongside dataset size and benchmark performance. Extended interaction logs contain sensitive mental health information.
Privacy-preserving protocols, potentially including minimization, consent-based collection, secure access controls, aggregation, or differential privacy where appropriate, are necessary complements. Mental health AI disproportionately reaches vulnerable users, those in crisis, with limited access to care, and young people, and the asymmetry between the cost of a missed failure and a false alarm should shape threshold choices and deployment pace. Anno-MI consists of transcripts of publicly available video demonstrations involving no identifiable individuals; no IRB approval was required.

\section{Conclusion}

We have argued that endpoint-style evidence can support invalid safety conclusions when it is used to certify temporal safety properties whose determining evidence has been discarded. Temporal Safety Non-Identifiability formalizes this problem: if two interactions with different safety status collapse to the same compressed representation, no evaluator using only that representation can recover the missing distinction.

SCOPE turns this observation into a reporting discipline. Evaluations should state the safety claim being made, the horizon over which it is assessed, the unit of analysis, the temporal determinants of the claim, the evidence preserved by the protocol, and the stronger claims that remain unsupported. SCOPE-MH instantiates this discipline for mental health AI through five temporal dimensions: longitudinal consistency, harm accumulation, intervention timing, recovery capability, and referral correctness.

The Anno-MI proof-of-concept illustrates the claim-evidence discipline rather than establishing a deployment-ready detector. The per-turn behavior baseline is the stronger classifier overall, while sustain-talk delta provides a minimal temporal signal that carries quality-relevant information, behaves as a selective audit trigger at an illustrative operating point, and reveals mechanisms and blind spots that the per-turn score itself does not represent. The case studies show both sides of the argument: temporal traces can reveal resistance accumulation, but simple temporal metrics can still miss stagnation and backsliding.

The question is not whether temporal evidence always improves prediction. The question is whether safety claims about temporal properties can be valid without preserving temporal evidence, and they cannot.

\bibliography{references}

@inproceedings{wu2022anno,
  title={Anno-{MI}: A dataset of expert-annotated counselling dialogues},
  author={Wu, Zixiu and Balloccu, Simone and Kumar, Vivek and Helaoui, Rim and Reiter, Ehud and Recupero, Diego Reforgiato and Riboni, Daniele},
  booktitle={ICASSP 2022-2022 IEEE International Conference on Acoustics, Speech and Signal Processing (ICASSP)},
  pages={6177--6181},
  year={2022},
  organization={IEEE}
}

@article{morrin2026journey,
  title={It Is the Journey, Not the Destination: Moving From End Points to Trajectories When Assessing Chatbot Mental Health Safety},
  author={Morrin, Hamilton and Yeung, Joshua Au and Agnew, Zarinah and {\O}stergaard, S{\o}ren Dinesen and Pollak, Thomas A},
  journal={JMIR Mental Health},
  volume={13},
  number={1},
  pages={e91454},
  year={2026},
  publisher={JMIR Publications Inc., Toronto, Canada}
}

@article{moyers2016motivational,
  title={The motivational interviewing treatment integrity code ({MITI} 4): Rationale, preliminary reliability and validity},
  author={Moyers, Theresa B and Rowell, Lauren N and Manuel, Jennifer K and Ernst, Denise and Houck, Jon M},
  journal={Journal of substance abuse treatment},
  volume={65},
  pages={36--42},
  year={2016},
  publisher={Elsevier}
}

@article{moreno2026umbrella,
  title={An umbrella review of psychological capacity and mental health trajectories across the life course},
  author={Moreno-Agostino, Dar{\'\i}o and Khan, Nusrat and De Rubeis, Vanessa and Jacob, Chandni Maria and Banati, Prerna and Sadana, Ritu and Prina, Matthew},
  journal={Nature Mental Health},
  pages={1--18},
  year={2026},
  publisher={Nature Publishing Group US New York}
}

@article{kandala2026echoguard,
  title={EchoGuard: An Agentic Framework with Knowledge-Graph Memory for Detecting Manipulative Communication in Longitudinal Dialogue},
  author={Kandala, Ratna and Manchanda, Niva and Moharir, Akshata Kishore and Kandala, Ananth},
  journal={arXiv preprint arXiv:2603.04815},
  year={2026}
}

@article{mcbain2025use,
  title={Use of generative {A}{I} for mental health advice among US adolescents and young adults},
  author={McBain, Ryan K and Bozick, Robert and Diliberti, Melissa and Zhang, Li Ang and Zhang, Fang and Burnett, Alyssa and Kofner, Aaron and Rader, Benjamin and Breslau, Joshua and Stein, Bradley D and Mehrotra, Ateev and Pines, Lori Uscher and Cantor, Jonathan and Yu, Hao},
  journal={JAMA Network Open},
  volume={8},
  number={11},
  pages={e2542281},
  year={2025},
  publisher={American Medical Association}
}

@article{rousmaniere2025large,
  title={Large language models as mental health resources: Patterns of use in the United States.},
  author={Rousmaniere, Tony and Zhang, Yimeng and Li, Xu and Shah, Siddharth},
  journal={Practice Innovations},
  year={2025},
  publisher={Educational Publishing Foundation}
}

@article{hua2025charting,
  title={Charting the evolution of artificial intelligence mental health chatbots from rule-based systems to large language models: A systematic review},
  author={Hua, Yining and Siddals, Steve and Ma, Zilin and Galatzer‐Levy, Isaac and Xia, Winna and Hau, Christine and Na, Hongbin and Flathers, Matthew and Linardon, Jake and Ayubcha, Cyrus and Torous, John},
  journal={World Psychiatry},
  volume={24},
  number={3},
  pages={383--394},
  year={2025},
  publisher={Wiley Online Library}
}

@article{zhang2026mhdash,
  title={{MHDash}: An Online Platform for Benchmarking Mental Health-Aware {AI} Assistants},
  author={Zhang, Yihe and Mohawk, Cheyenne N and Han, Kaiying and Tida, Vijay Srinivas and Li, Manyu and Hei, Xiali},
  journal={arXiv preprint arXiv:2602.00353},
  year={2026}
}

@article{pombal2025mindeval,
  title={Mind{E}val: Benchmarking Language Models on Multi-turn Mental Health Support},
  author={Pombal, Jos{\'e} and D'Eon, Maya and Guerreiro, Nuno M and Martins, Pedro Henrique and Farinhas, Ant{\'o}nio and Rei, Ricardo},
  journal={arXiv preprint arXiv:2511.18491},
  year={2025}
}

@article{tanana2016comparison,
  title={A comparison of natural language processing methods for automated coding of motivational interviewing},
  author={Tanana, Michael and Hallgren, Kevin A and Imel, Zac E and Atkins, David C and Srikumar, Vivek},
  journal={Journal of substance abuse treatment},
  volume={65},
  pages={43--50},
  year={2016},
  publisher={Elsevier}
}

@article{fitzpatrick2017delivering,
  title={Delivering cognitive behavior therapy to young adults with symptoms of depression and anxiety using a fully automated conversational agent ({W}oebot): {A} randomized controlled trial},
  author={Fitzpatrick, Kathleen Kara and Darcy, Alison and Vierhile, Molly},
  journal={JMIR mental health},
  volume={4},
  number={2},
  pages={e7785},
  year={2017},
  publisher={JMIR Publications Inc., Toronto, Canada}
}

@article{li2023systematic,
  title={Systematic review and meta-analysis of {AI}-based conversational agents for promoting mental health and well-being},
  author={Li, Han and Zhang, Renwen and Lee, Yi-Chieh and Kraut, Robert E and Mohr, David C},
  journal={NPJ Digital Medicine},
  volume={6},
  number={1},
  pages={236},
  year={2023},
  publisher={Nature Publishing Group UK London}
}

@article{hendrycks2020measuring,
  title={Measuring massive multitask language understanding},
  author={Hendrycks, Dan and Burns, Collin and Basart, Steven and Zou, Andy and Mazeika, Mantas and Song, Dawn and Steinhardt, Jacob},
  journal={arXiv preprint arXiv:2009.03300},
  year={2020}
}

@misc{liang2023holisticevaluationlanguagemodels,
      title={Holistic Evaluation of Language Models}, 
      author={Percy Liang and Rishi Bommasani and Tony Lee and Dimitris Tsipras and Dilara Soylu and Michihiro Yasunaga and Yian Zhang and Deepak Narayanan and Yuhuai Wu and Ananya Kumar and Benjamin Newman and Binhang Yuan and Bobby Yan and Ce Zhang and Christian Cosgrove and Christopher D. Manning and Christopher Ré and Diana Acosta-Navas and Drew A. Hudson and Eric Zelikman and Esin Durmus and Faisal Ladhak and Frieda Rong and Hongyu Ren and Huaxiu Yao and Jue Wang and Keshav Santhanam and Laurel Orr and Lucia Zheng and Mert Yuksekgonul and Mirac Suzgun and Nathan Kim and Neel Guha and Niladri Chatterji and Omar Khattab and Peter Henderson and Qian Huang and Ryan Chi and Sang Michael Xie and Shibani Santurkar and Surya Ganguli and Tatsunori Hashimoto and Thomas Icard and Tianyi Zhang and Vishrav Chaudhary and William Wang and Xuechen Li and Yifan Mai and Yuhui Zhang and Yuta Koreeda},
      year={2023},
      eprint={2211.09110},
      archivePrefix={arXiv},
      primaryClass={cs.CL},
      url={https://arxiv.org/abs/2211.09110}, 
}

@misc{park2025buildingtrustmentalhealth,
      title={Building Trust in Mental Health Chatbots: Safety Metrics and {LLM}-Based Evaluation Tools}, 
      author={Jung In Park and Mahyar Abbasian and Iman Azimi and Dawn T. Bounds and Angela Jun and Jaesu Han and Robert M. McCarron and Jessica Borelli and Parmida Safavi and Sanaz Mirbaha and Jia Li and Mona Mahmoudi and Carmen Wiedenhoeft and Amir M. Rahmani},
      year={2025},
      eprint={2408.04650},
      archivePrefix={arXiv},
      primaryClass={cs.CL},
      url={https://arxiv.org/abs/2408.04650}, 
}

@article{li2025counselbench,
  title={{CounselBench}: A large-scale expert evaluation and adversarial benchmarking of large language models in mental health question answering},
  author={Li, Yahan and Yao, Jifan and Bunyi, John Bosco S and Frank, Adam C and Hwang, Angel Hsing-Chi and Liu, Ruishan},
  journal={arXiv preprint arXiv:2506.08584},
  year={2025}
}

@article{lee2026mhsafeeval,
  title={{MHSafeEval}: Role-Aware Interaction-Level Evaluation of Mental Health Safety in Large Language Models},
  author={Lee, Suhyun and Achananuparp, Palakorn and Yadav, Neemesh and Lim, Ee-Peng and Deng, Yang},
  journal={arXiv preprint arXiv:2604.17730},
  year={2026}
}

@article{Stringer2026,
  author  = {Stringer, Heather},
  title   = {{AI}, neuroscience, and data are fueling personalized mental health care},
  journal = {Monitor on Psychology},
  year    = {2026},
  month   = {January/February},
  volume  = {57},
  number  = {1},
  pages   = {56},
  url     = {https://www.apa.org/monitor/2026/01-02/trends-personalized-mental-health-care}
}

@article{ngabo2025prognostic,
  title={A Prognostic Theory of Treatment Response for Major Depressive Disorder: A Dynamic Systems Framework for Forecasting Clinical Trajectories},
  author={Ngabo-Woods, Harold and Dunai, Larisa and Verd{\'u}, Isabel Segu{\'\i}},
  journal={Applied Sciences},
  volume={15},
  number={23},
  pages={12524},
  year={2025},
  publisher={MDPI}
}

@article{bosl2025dynamical,
  title={A dynamical systems framework for precision psychiatry},
  author={Bosl, William J and Enlow, Michelle Bosquet and Nelson, Charles A},
  journal={npj Digital Medicine},
  volume={8},
  number={1},
  pages={586},
  year={2025},
  publisher={Nature Publishing Group UK London}
}

\newpage
\appendix

\appendix

\section{Temporal Safety Non-Identifiability: Formal Details}
\label{app:tsni}

The main paper states Temporal Safety Non-Identifiability in compact form. Here we provide the full definitions and proof details.

Let \(\mathcal{T}\) denote the set of possible interaction trajectories. A trajectory \(\tau \in \mathcal{T}\) is an ordered sequence of user states, system actions, contextual information, and risk-relevant variables over an interaction horizon \(H\). Let \(S(\tau)\) denote the safety status of the trajectory with respect to a specified safety predicate. In practice, \(S\) may be binary, graded, probabilistic, multidimensional, or expert-adjudicated; the simplified notation is used only to state the identifiability problem. In probabilistic settings, \(S(\tau)\) can be read as the risk distribution, expected risk, or calibrated probability assigned to \(\tau\). The non-identifiability claim applies whenever the safety-relevant quantity returned by \(S\) differs across two trajectories collapsed by the evaluation representation.

An endpoint-style evaluation protocol observes the trajectory only through a compression function
\[
\phi: \mathcal{T} \rightarrow \mathcal{Z},
\]
where \(\phi(\tau)\) may include terminal, aggregate, or instance-level quantities such as final symptom change, average response score, satisfaction, engagement, adverse-event count, final answer, aggregate transcript score, event count, or benchmark performance. Any evaluator that relies only on this compressed representation can be written as
\[
E_{\mathrm{end}}(\tau) = g(\phi(\tau)),
\]
for some scoring or decision rule \(g: \mathcal{Z} \rightarrow \mathcal{Y}\).

\paragraph{Definition 1: Temporal Safety Non-Identifiability.}
An endpoint-style evaluation protocol exhibits \emph{Temporal Safety Non-Identifiability} for a safety predicate \(S\) when there exist two trajectories \(\tau, \tau' \in \mathcal{T}\) such that
\[
\phi(\tau) = \phi(\tau')
\]
but
\[
S(\tau) \neq S(\tau').
\]
In such a case, the evaluator receives the same compressed representation for two trajectories with different safety status.

\paragraph{Proposition 1: Compression-Induced Non-Identifiability of Temporal Safety.}
Let \(S\) be a safety predicate over trajectories, and let \(\phi: \mathcal{T} \rightarrow \mathcal{Z}\) be the representation observed by an evaluation protocol. If there exist trajectories \(\tau, \tau' \in \mathcal{T}\) such that \(\phi(\tau)=\phi(\tau')\) but \(S(\tau) \neq S(\tau')\), then no deterministic evaluator \(g(\phi(\tau))\) can correctly determine \(S\) for all trajectories in \(\mathcal{T}\). Likewise, no randomized evaluator whose output distribution depends only on \(\phi(\tau)\) can guarantee correct determination for both trajectories.

Equivalently, \(S\) is identifiable from \(\phi\) only if \(S\) is constant on the equivalence classes induced by \(\phi\). That is, all trajectories mapped to the same compressed representation must have the same safety status. If this condition fails, no downstream scoring rule can recover the missing distinction.

\paragraph{Proof sketch.}
If \(\phi(\tau)=\phi(\tau')\), then every deterministic evaluator that depends only on \(\phi\) receives the same input for \(\tau\) and \(\tau'\). It must therefore assign the same score, label, or decision to both trajectories. If \(S(\tau) \neq S(\tau')\), then at least one assignment must be wrong. A randomized evaluator whose output distribution depends only on \(\phi\) faces the same problem: because the induced output distribution is identical for both trajectories, it cannot guarantee correct determination for two trajectories with different safety status. The failure is not a matter of sample size, calibration, annotation quality, or evaluator sophistication. It is a consequence of information loss.

For example, two systems may produce the same final symptom score, the same average response-quality rating, and the same number of crisis-resource mentions. One escalates at the first credible sign of imminent risk. The other escalates only after several turns of reinforcement, delay, and user disengagement. If the evaluator observes only the compressed representation, these trajectories are indistinguishable even though their safety status differs.

This result is deliberately modest. It does not show that endpoint measures are useless, nor does it show that every safety failure requires long-term evaluation. It shows something narrower and more important for evaluation design: endpoint-style evaluation cannot certify temporal safety properties when the compression function removes the temporal features on which safety depends.


\subsection{Definitions and Scope}
\label{app:definitions}

A trajectory is an ordered sequence of interaction states over an interaction horizon \(H\):
\[
\tau_{1:H} = \{(x_t, a_t, h_t, r_t)\}_{t=1}^{H}.
\]

Here, \(x_t\) denotes the user's observable input or expressed state at turn \(t\), \(a_t\) denotes the system action or response, \(h_t\) denotes the available interaction history and contextual information, and \(r_t\) denotes a risk-relevant state, whether observed or inferred. Such states may include distress, suicidal ideation, self-harm risk, dependency, disengagement, impaired reality testing, crisis severity, or need for referral. A trajectory may also include terminal outcomes, but its defining feature is ordering: the same set of turns can represent different safety objects when arranged differently.

Endpoint evaluation refers to evaluation based on terminal or post-hoc outcomes, including symptom reduction, satisfaction, adherence, disengagement, or adverse-event summaries. Instance-based evaluation refers to evaluation based on isolated prompts, individual responses, short exchanges, or independently scored benchmark items. These are distinct evaluation forms. We use endpoint-style evaluation as a broader term for any protocol that bases its primary safety judgment on terminal, aggregate, or temporally compressed quantities. Endpoint and instance-based evaluation become endpoint-style, in the sense used here, when they discard the temporal structure needed to assess the safety property being claimed.

The interaction horizon is the temporal span over which system behavior may influence user state and safety-relevant outcomes. It may be measured in turns, sessions, days, weeks, or clinically meaningful phases of use. A one-turn horizon may detect overtly unsafe responses. A session-level horizon may detect missed escalation, inconsistency, or delayed repair. A longitudinal horizon may be needed to detect dependency, disengagement, habituation, delayed deterioration, or cumulative reinforcement of maladaptive beliefs.

By temporal-evidence evaluation, we mean evaluation that treats the ordered interaction as part of the object being assessed. Such evaluation asks whether the system's behavior remains safe as the user's state evolves, rather than inferring safety solely from final outcomes or isolated responses.

These definitions establish the paper's central methodological claim. Endpoint outcomes remain necessary for evaluating whether users improve, and instance-based benchmarks remain useful for assessing local response quality. But neither can certify temporal safety unless the evaluation preserves the temporal structure on which that safety depends. For mental health AI, where harm may arise through delay, accumulation, inconsistency, failed recovery, or mistimed referral, the trajectory is not supporting context. It is part of the safety object.

This claim motivates SCOPE, the reporting standard developed in Section~\ref{sec:scope}, and its mental-health instantiation, SCOPE-MH. SCOPE-MH operationalizes the principle that safety claims must be constrained by the temporal structure preserved in evaluation. It asks whether a system remains safe across five temporal dimensions: longitudinal consistency, harm accumulation, intervention timing, recovery capability, and referral correctness.

\vspace{- 10 mm}

\newpage
\section{How Claim-Evidence Mismatch Produces False Assurance}
\label{app:false-assurance}


Table~\ref{tab:false-safety-assurance-app} expands the three practical failure modes summarized in the main text. These are not independent objections to endpoint-style evidence. They are recurring ways in which an evaluation measures one object while licensing a stronger temporal safety claim about another.

\begin{table}[H]
\centering
\caption{Three failure modes through which endpoint-style evaluation can produce false safety assurance.}
\label{tab:false-safety-assurance-app}
\small
\setlength{\tabcolsep}{3.5pt}
\renewcommand{\arraystretch}{1.12}
\begin{tabular}{
@{}
>{\raggedright\arraybackslash}p{0.22\linewidth}
>{\raggedright\arraybackslash}p{0.25\linewidth}
>{\raggedright\arraybackslash}p{0.23\linewidth}
>{\raggedright\arraybackslash}p{0.22\linewidth}
@{}}
\toprule
\textbf{Failure mode}
  & \textbf{What evaluation measures}
  & \textbf{What it misses}
  & \textbf{False safety conclusion} \\
\midrule
\textbf{Unit-of-analysis mismatch}
  & Isolated prompts, individual responses, or short exchanges
  & System behavior as user state evolves
  & Safe responses imply a safe interactive system. \\
\addlinespace[0.35em]
\textbf{Temporal evidence loss}
  & Final answer, average score, event count, or compressed transcript
  & Timing, escalation latency, accumulation, adaptation, and recovery
  & The final or average score demonstrates safe behavior over time. \\
\addlinespace[0.35em]
\textbf{Outcome attribution gap}
  & Symptom change, satisfaction, engagement, referral uptake, disengagement, or adverse events
  & Whether the system caused or safely supported the outcome
  & Favorable outcomes imply safe system behavior. \\
\bottomrule
\end{tabular}
\end{table}

A response-level evaluation can support claims about local behavior under specified prompts, but not claims about whether the system remains safe as user state evolves. A multi-turn evaluation can still fail if it scores only the final response, averages quality across turns, counts whether an event occurred without recording when it occurred, or treats each turn as independently scorable. Endpoint outcomes such as symptom reduction, satisfaction, and engagement remain essential, but they do not automatically show that the system behaved safely while the user reached the outcome.

\newpage
\section{Anno-MI Dataset and Preprocessing}
\label{app:dataset}

Table~\ref{tab:dataset_stats} summarises the key statistics of the Anno-MI dataset. The full dataset contains 133 conversations. After excluding two conversations with fewer than four labeled client turns, the analysis uses 131 conversations: 110 high-quality and 21 low-quality. Both excluded conversations are low-quality conversations, which accounts for the difference between the 23 low-quality conversations in the full dataset and the 21 low-quality conversations used in analysis.

\begin{table}[H]
\centering
\caption{Anno-MI dataset statistics.}
\label{tab:dataset_stats}
\small
\begin{tabular}{lc}
\toprule
\textbf{Property} & \textbf{Value} \\
\midrule
Total conversations                  & 133           \\
\quad High quality                   & 110 (82.7\%)  \\
\quad Low quality                    & 23 (17.3\%)   \\
\midrule
Conversations used in analysis       & 131           \\
\quad High quality used              & 110           \\
\quad Low quality used               & 21            \\
\quad Excluded (fewer than 4 labeled client turns) & 2 \\
\midrule
\multicolumn{2}{l}{\textit{Turns per conversation (all 133):}} \\
\quad Mean                           & 72.9          \\
\quad Std deviation                  & 84.3          \\
\quad Minimum                        & 6             \\
\quad 25th percentile                & 26            \\
\quad Median                         & 47            \\
\quad 75th percentile                & 79            \\
\quad Maximum                        & 598           \\
\bottomrule
\end{tabular}
\end{table}

Utterances are sorted by temporal order and separated into therapist and client turns. Therapist turns are used for the per-turn behavior baseline. Client turns with talk-type labels are used for temporal client-language metrics. Conversations with fewer than four labeled client turns are excluded because midpoint-split temporal metrics become unstable with fewer than two labeled client turns per half.

\newpage
\section{Metric Definitions}
\label{app:metrics}

We compute two families of scores for every conversation: a per-turn behavior baseline and temporal client-language metrics. Both use only information available in the annotated conversation log.

\paragraph{Per-turn behavior baseline.}
The per-turn behavior baseline is the fraction of therapist turns involving reflection or questioning, the behaviors most associated with high-quality motivational interviewing practice:
\begin{equation}
  \mathrm{PT} =
    \frac{\sum_{t \in \mathcal{T}_{\text{therapist}}}
          \mathbf{1}[\mathrm{beh}_t \in \{\mathrm{reflection}, \mathrm{question}\}]}
         {|\mathcal{T}_{\text{therapist}}|}.
  \label{eq:perturn}
\end{equation}
This score represents the quality of individual therapist responses aggregated over the conversation. It does not represent how client language changes over time.

\paragraph{Sustain-talk delta.}
The primary temporal metric is sustain-talk delta. We split each conversation at the midpoint of labeled client utterances and compute the change in client resistance talk between halves:
\begin{equation}
  \Delta_{\text{sustain}} =
    \underbrace{\frac{1}{\lfloor n/2 \rfloor}
    \sum_{i \leq n/2} \mathbf{1}[\tau_i = \mathrm{sustain}]}_{\text{first-half sustain ratio}}
    \;-\;
    \underbrace{\frac{1}{n - \lfloor n/2 \rfloor}
    \sum_{i > n/2}  \mathbf{1}[\tau_i = \mathrm{sustain}]}_{\text{second-half sustain ratio}} .
  \label{eq:delta}
\end{equation}
A positive \(\Delta_{\text{sustain}}\) indicates decreasing resistance across the conversation. A negative value indicates increasing resistance.

\paragraph{Temporal score.}
The temporal score combines change-talk increase and sustain-talk decrease:
\begin{equation}
  \mathrm{TEMP} =
  \underbrace{(\bar{c}_2 - \bar{c}_1)}_{\text{change-talk increase}}
  +
  \underbrace{(\bar{s}_1 - \bar{s}_2)}_{\text{sustain-talk decrease}},
  \label{eq:temporal-score}
\end{equation}
where \(\bar{c}_k\) and \(\bar{s}_k\) are change-talk and sustain-talk ratios in conversation half \(k\), respectively.

\paragraph{Timing and persistence metrics.}
We also compute timing and persistence metrics motivated by SCOPE-MH's intervention timing and recovery capability dimensions.

\begin{itemize}
  \item \textbf{First sustain dominance} (\(T_{\mathrm{dom}}\)): the raw client-turn index at which sustain-talk first dominates a three-turn window. Higher values indicate later onset of resistance.
  \item \textbf{Dominance normalized} (\(T_{\mathrm{norm}}\)): \(T_{\mathrm{dom}}\) divided by the number of labeled client turns. This maps first sustain dominance to the conversation's relative temporal position.
  \item \textbf{Maximum sustain streak} (\(S_{\max}\)): the longest consecutive run of sustain-talk turns, used as a proxy for unresolved resistance that was not interrupted or repaired.
\end{itemize}

If no sustain-dominance event occurs, the event is treated as right-censored at the end of the labeled client-turn sequence for descriptive timing summaries. This convention preserves the interpretation that larger values correspond to later onset of resistance.

\paragraph{Early-warning metric.}
The early-warning signal differs from full-conversation sustain-talk delta because, at the midpoint, only the first half of the conversation is observed. We compute:
\begin{equation}
  EW_{\Delta} = \bar{c}_1 - \bar{s}_1,
  \label{eq:early-warning}
\end{equation}
where \(\bar{c}_1\) and \(\bar{s}_1\) are the change-talk and sustain-talk ratios in the first half. Lower values indicate that resistance already outweighs motivation at midpoint. We flag early risk when
\[
EW_{\Delta} < \theta_{\mathrm{ew}},
\]
with \(\theta_{\mathrm{ew}}=-0.15\), selected by F1 maximization on the early-warning task. This sign convention keeps negative values aligned with temporal deterioration.

\newpage
\section{Full Empirical Results}
\label{app:full-results}

\subsection{Full Metric Comparison}
\label{app:full-metrics}

Table~\ref{tab:full-metrics} reports all six metrics computed across the 131 analyzed conversations.

\begin{table}[H]
\centering
\caption{Metric comparison across 131 Anno-MI conversations. Two-sample t-tests are reported with Cohen's \(d\) using pooled standard deviation.}
\label{tab:full-metrics}
\small
\setlength{\tabcolsep}{4pt}
\renewcommand{\arraystretch}{1.08}
\begin{tabular}{lcccccc}
\toprule
\textbf{Metric} & \textbf{Type} & \textbf{High} & \textbf{Low} & \textbf{Diff} & \textbf{\(p\)} & \textbf{Cohen's \(d\)} \\
\midrule
Per-turn process score & Per-turn baseline & 0.598 & 0.292 & +0.306 & \(<0.001\) & 1.599*** \\
Sustain-talk delta (\(\Delta_s\)) & Temporal direction & 0.062 & -0.053 & +0.115 & 0.037 & 0.462* \\
Temporal score & Temporal direction & 0.231 & 0.091 & +0.140 & 0.137 & 0.353 \\
Dominance normalized (\(T_{\mathrm{norm}}\)) & Timing & 0.425 & 0.341 & +0.083 & 0.397 & 0.211 \\
First sustain dominance (\(T_{\mathrm{dom}}\)) & Timing & 8.623 & 5.941 & +2.682 & 0.541 & 0.177 \\
Max sustain streak (\(S_{\max}\)) & Persistence & 1.727 & 2.190 & -0.463 & 0.252 & -0.274 \\
\bottomrule
\end{tabular}

\vspace{0.35em}
\footnotesize{*** \(p<0.001\); * \(p<0.05\).}
\end{table}

The per-turn behavior baseline is the strongest classifier in this dataset. Sustain-talk delta is the strongest temporal signal and separates high- and low-quality conversations under a parametric test, but the corresponding Mann--Whitney test is not significant (Appendix~\ref{app:nonparametric}). We therefore treat sustain-talk delta as quality-relevant but preliminary.

\subsection{Aggregate Temporal Pattern}
\label{app:aggregate-pattern}

\begin{figure}[H]
  \centering
  \includegraphics[width=0.86\linewidth]{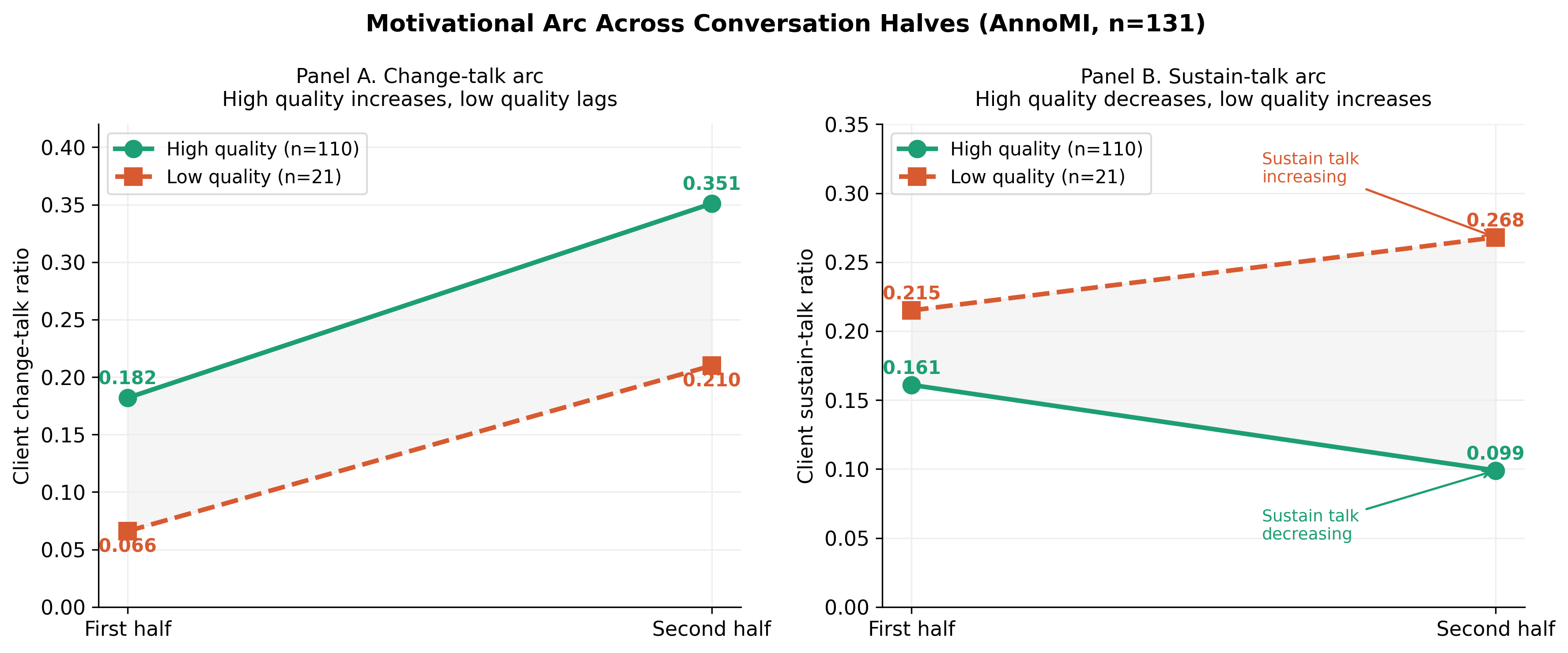}
  \caption{Client-language arcs across conversation halves by expert quality label. High-quality conversations show stronger movement toward change-talk and decreasing sustain-talk, while low-quality conversations show weaker change-talk growth and increasing sustain-talk.}
  \label{fig:change-arc-app}
\end{figure}

Figure~\ref{fig:change-arc-app} shows aggregate change-talk and sustain-talk patterns across conversation halves. High-quality conversations show increasing change-talk and decreasing sustain-talk across halves, while low-quality conversations show weaker change-talk growth and increasing sustain-talk.

\subsection{Full-Conversation Detection}
\label{app:full-detection}

For full-conversation detection, the temporal signal flags a conversation as low quality when \(\Delta_s < \theta\), with \(\theta=-0.10\) selected by F1 maximization across thresholds from \(-0.05\) to \(-0.50\). The per-turn behavior baseline flags a conversation as low quality when \(\mathrm{PT}<\theta_{\mathrm{PT}}\), where \(\theta_{\mathrm{PT}}\) is defined in Appendix~\ref{app:metrics}. Table~\ref{tab:full-detection} reports the resulting operating point. Because thresholds are selected or reported on the same dataset used for evaluation, thresholded detection should be interpreted as illustrative rather than deployment-validated.

\begin{table}[H]
\centering
\caption{Full-conversation detection of 21 low-quality conversations.}
\label{tab:full-detection}
\small
\setlength{\tabcolsep}{5pt}
\renewcommand{\arraystretch}{1.08}
\begin{tabular}{lcccc}
\toprule
\textbf{Method} & \textbf{Detected} & \textbf{Detection rate} & \textbf{False alarms} & \textbf{False alarm rate} \\
\midrule
Per-turn behavior baseline & 19/21 & 90\% & 55/110 & 50\% \\
Temporal signal & 7/21 & 33\% & 13/110 & 12\% \\
Combined (either) & 19/21 & 90\% & 55/110 & 50\% \\
\bottomrule
\end{tabular}
\end{table}

The temporal signal does not add full-conversation recall beyond the per-turn behavior baseline at this operating point. Its value is selectivity: it flags fewer conversations and produces fewer false alarms among high-quality conversations.

\subsection{Midpoint Early Warning}
\label{app:midpoint}

For midpoint early warning, only the first half of the conversation is used. The temporal early-warning score \(EW_{\Delta}\) is defined in Appendix~\ref{app:metrics}. The threshold \(\theta_{\mathrm{ew}}=-0.15\) is selected by F1 maximization on the early-warning task and should be treated as illustrative. Table~\ref{tab:midpoint-detection} reports the resulting midpoint operating point.

\begin{table}[H]
\centering
\caption{Early-warning detection at conversation midpoint.}
\label{tab:midpoint-detection}
\small
\setlength{\tabcolsep}{5pt}
\renewcommand{\arraystretch}{1.08}
\begin{tabular}{lcccc}
\toprule
\textbf{Method} & \textbf{Detected} & \textbf{Detection rate} & \textbf{False alarms} & \textbf{False alarm rate} \\
\midrule
Per-turn behavior baseline & 13/21 & 62\% & 22/110 & 20\% \\
Temporal signal & 9/21 & 43\% & 25/110 & 23\% \\
Combined (either) & 14/21 & 67\% & 35/110 & 32\% \\
\bottomrule
\end{tabular}
\end{table}

At midpoint, the temporal signal uniquely catches one low-quality conversation missed by the per-turn behavior baseline, but the combined rule also increases false alarms. This result should be interpreted as limited complementary early-warning coverage, not as a validated intervention policy.

\newpage
\section{Case Details and SCOPE-MH Error Analysis}
\label{app:cases}

The main text discusses Conversation~27 as an example of mechanism visibility: both evidence sources flag the conversation as low quality, but the temporal trace shows the resistance accumulation mechanism. Table~\ref{tab:case-error-app} summarizes this case together with two low-quality conversations that both the per-turn baseline and the simple temporal signal miss.

\begin{table}[H]
\centering
\caption{SCOPE-MH error analysis of three low-quality Anno-MI conversations. Temporal evidence can reveal mechanisms, but simple temporal metrics do not capture all temporal failure modes.}
\label{tab:case-error-app}
\small
\setlength{\tabcolsep}{3.5pt}
\renewcommand{\arraystretch}{1.12}
\begin{tabular}{
@{}
>{\raggedright\arraybackslash}p{0.09\linewidth}
>{\raggedright\arraybackslash}p{0.14\linewidth}
>{\raggedright\arraybackslash}p{0.16\linewidth}
>{\raggedright\arraybackslash}p{0.16\linewidth}
>{\raggedright\arraybackslash}p{0.37\linewidth}
@{}}
\toprule
\textbf{Case}
& \textbf{Expert label}
& \textbf{Per-turn baseline}
& \textbf{Temporal signal}
& \textbf{SCOPE implication} \\
\midrule
27
& Low
& Fail
& Fail
& Correct verdict does not imply adequate evidence: the temporal trace reveals resistance accumulation after a directive intervention. \\
\addlinespace[0.35em]
53
& Low
& Pass
& Pass
& Simple sustain-talk delta misses stagnation: a long neutral plateau can indicate lack of therapeutic progress without overt resistance growth. \\
\addlinespace[0.35em]
131
& Low
& Pass
& Pass
& Simple sustain-talk delta misses backsliding: change-talk can deteriorate even when sustain-talk does not increase. \\
\bottomrule
\end{tabular}
\end{table}

\subsection{Conversation 53: Stagnation Without Overt Resistance Growth}
\label{app:conv53}
\vspace{- 3.15 mm}

Conversation~53 is an alcohol-reduction conversation with 108 turns. It scores above threshold on both evidence sources: the per-turn behavior baseline is \(0.630\), and sustain-talk delta is \(\Delta_s = 0.000\). The talk-type sequence reveals why both methods miss the expert low-quality label: 37 consecutive neutral turns are followed by a burst of change-talk in the second half, producing an arc that looks acceptable by the current metrics.

The likely failure is stagnation rather than overt resistance growth. The therapist does not appear to build sustained motivational movement across a long neutral period, but the midpoint-split sustain-talk metric does not represent stagnation. This case motivates a separate stagnation-duration measure: the longest stretch of neutral or non-progressive client language before meaningful change-talk emerges.

\vspace{- 1 mm}

\subsection{Conversation 131: Backsliding}
\label{app:conv131}

\vspace{-1 mm}

Conversation~131 is a smoking-cessation conversation with 91 turns. It illustrates a different temporal failure: backsliding. The client expresses strong motivation early, with 8 change-talk turns in the first half, but this drops to 3 change-talk turns in the second half. Sustain-talk also decreases, so the simple sustain-talk delta remains acceptable \((\Delta_s = +0.093)\), and both the per-turn behavior baseline and the temporal signal pass the conversation.

The expert label is low quality. The missed temporal pattern is not growing resistance, but loss of earlier motivation. This case motivates a backsliding index that tracks decline in change-talk after an early high-change phase, regardless of whether sustain-talk also decreases.

\subsection{Implication for SCOPE-MH}
\label{app:case-implication}

These cases show why SCOPE-MH is not a single-metric proposal. Sustain-talk delta preserves evidence about one temporal property: resistance accumulation. It does not preserve all temporal failure modes. Stagnation, backsliding, delayed referral, failure to recover, or longitudinal inconsistency require different signals. The role of SCOPE-MH is to state which evidence a metric preserves, which claims it can support, and which claims remain unidentified.

\newpage
\section{Additional Statistical Analyses}
\label{app:stats}

\subsection{ROC and Precision-Recall Curves}
\label{app:roc-pr}

\begin{figure}[H]
  \centering
  \includegraphics[width=\linewidth]{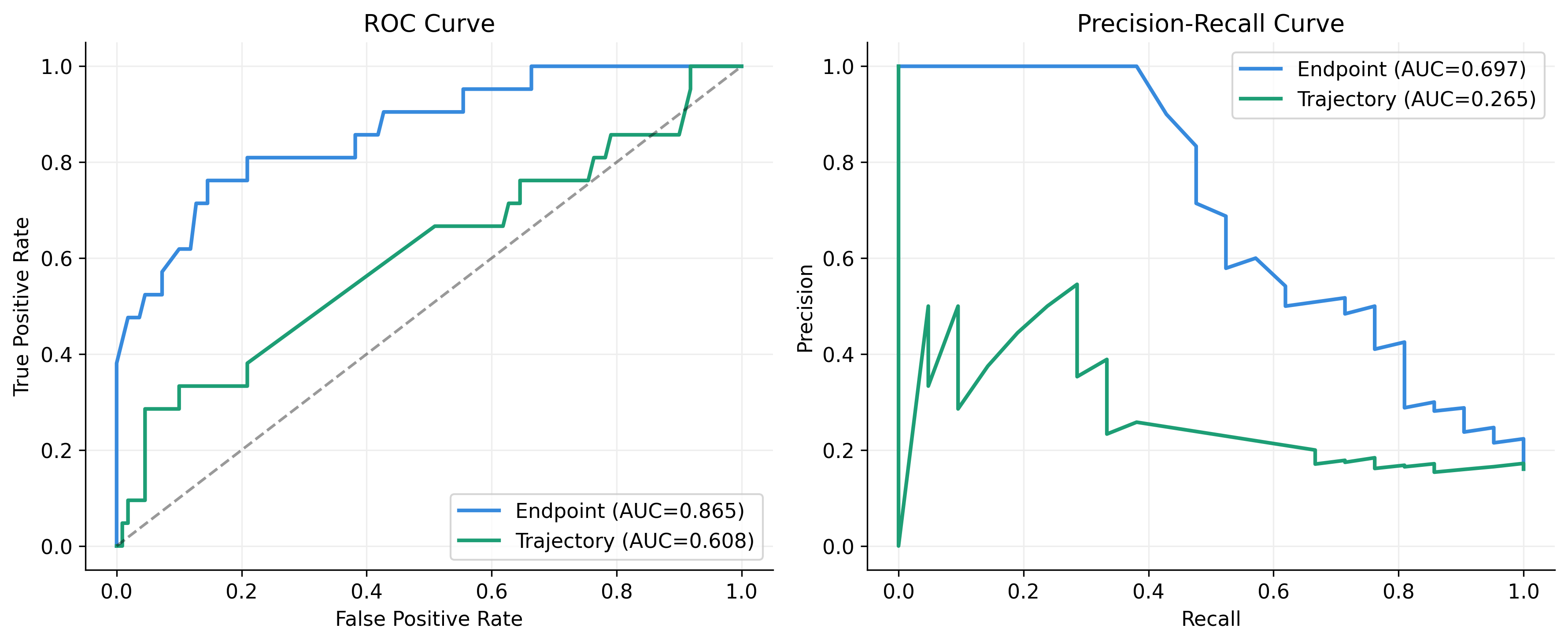}
  \caption{ROC and precision-recall curves on Anno-MI (\(n=131\)). In the plot legend, ``Endpoint'' refers to the per-turn behavior baseline, and ``Trajectory'' refers to the temporal signal. The per-turn behavior baseline achieves higher threshold-free performance (ROC AUC \(=0.865\), PR AUC \(=0.697\)) than the temporal signal (ROC AUC \(=0.608\), PR AUC \(=0.265\)), while the temporal signal is evaluated in the paper as a selective audit trigger rather than a standalone classifier.}
  \label{fig:roc}
\end{figure}

Figure~\ref{fig:roc} presents ROC and precision-recall curves across detection thresholds. The figure uses earlier shorthand labels: ``Endpoint'' denotes the per-turn behavior baseline, and ``Trajectory'' denotes the temporal signal. The per-turn behavior baseline achieves substantially higher AUC on both curves (ROC: 0.865 vs.\ 0.608; PR: 0.697 vs.\ 0.265), consistent with its stronger separation in Table~\ref{tab:full-metrics}. These curves confirm that the per-turn behavior baseline remains the stronger classifier overall, while the temporal signal behaves as a more selective audit trigger at the selected full-conversation threshold.

\subsection{Bootstrap Confidence Intervals}
\label{app:bootstrap}

Bootstrap confidence intervals with 1000 resamples are reported for the selected full-conversation threshold \(\theta=-0.10\). The observed rates are reported first; confidence intervals are bootstrap percentile intervals.

\begin{itemize}
  \item Observed detection rate: \(0.33\); bootstrap 95\% CI \([0.14,\ 0.52]\).
  \item Observed false alarm rate: \(0.12\); bootstrap 95\% CI \([0.05,\ 0.16]\).
\end{itemize}

The wide confidence interval on detection reflects the small low-quality sample (\(n=21\)). These estimates should therefore be treated as proof-of-concept operating characteristics rather than validated deployment performance.

\subsection{Nonparametric Tests}
\label{app:nonparametric}

Given the small low-quality sample and potential non-normality, Table~\ref{tab:nonparam} reports Mann--Whitney U tests alongside the parametric t-tests.

\begin{table}[H]
\centering
\caption{Mann--Whitney U test results for all six metrics.}
\label{tab:nonparam}
\small
\setlength{\tabcolsep}{5pt}
\renewcommand{\arraystretch}{1.08}
\begin{tabular}{lccc}
\toprule
\textbf{Metric} & \textbf{U statistic}
  & \textbf{\(p\) (t-test)} & \textbf{\(p\) (Mann--Whitney)} \\
\midrule
Per-turn process score              & 1997.5 & \(<0.001\) & \(<0.001\) \\
Sustain-talk delta (\(\Delta_s\))   & 1405.5 & 0.037 & 0.112 \\
Temporal score                      & 1416.0 & 0.137 & 0.102 \\
Dominance normalized                & 1263.0 & 0.397 & 0.493 \\
First sustain dominance             & 750.0  & 0.541 & 0.343 \\
Max sustain streak                  & 949.0  & 0.252 & 0.187 \\
\bottomrule
\end{tabular}
\end{table}

The discrepancy for sustain-talk delta (\(p=0.037\) parametric vs.\ \(p=0.112\) nonparametric) reflects the small sample size and potential distributional assumptions of the t-test. We therefore treat sustain-talk delta as a quality-relevant but preliminary temporal signal requiring replication on larger datasets.

\subsection{Threshold Sensitivity Analysis}
\label{app:threshold-sensitivity}

\begin{figure}[H]
  \centering
  \includegraphics[width=0.80\linewidth]{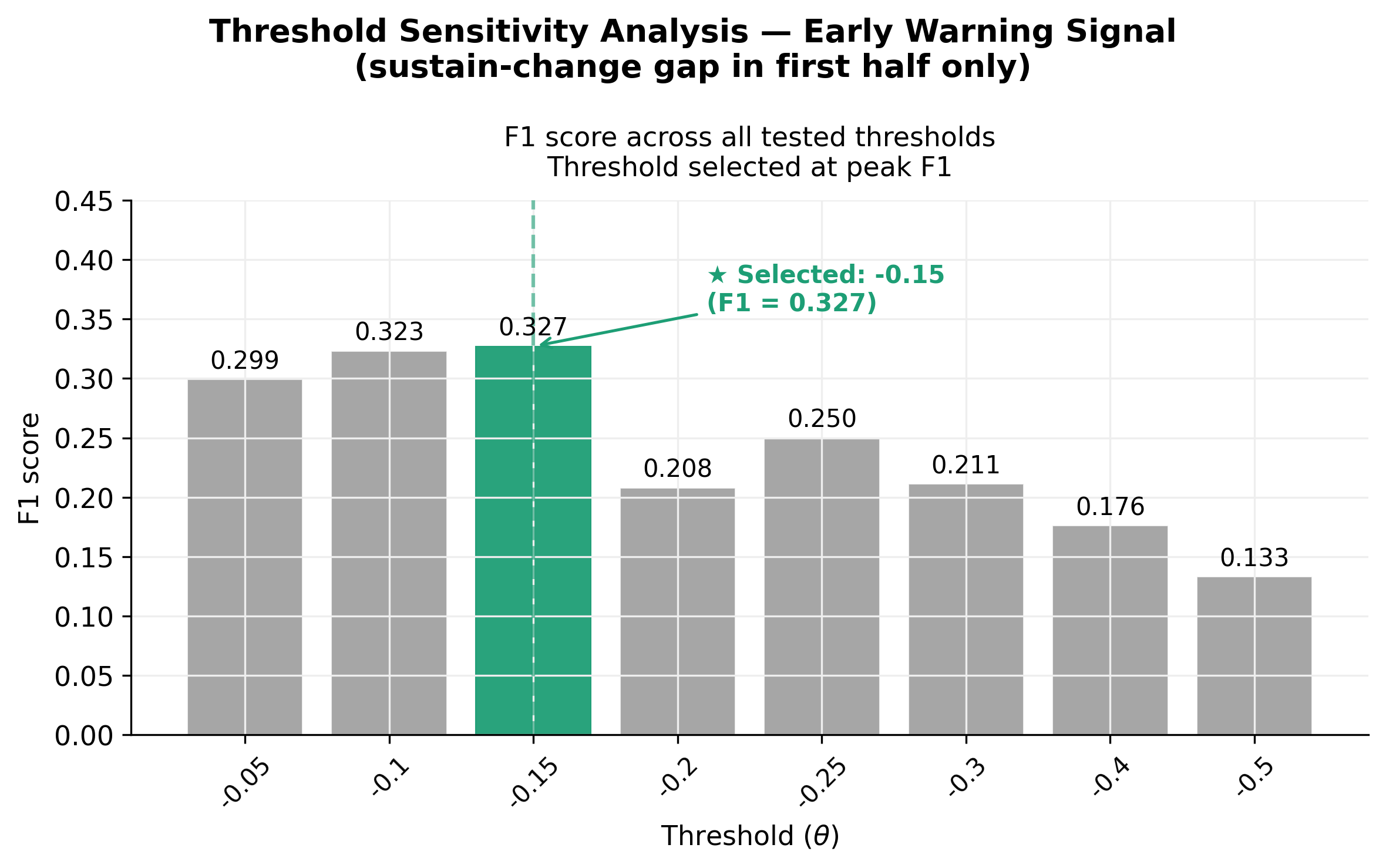}
  \caption{Early-warning threshold sensitivity analysis. The plot was generated using earlier shorthand for the midpoint sustain-change gap. In the paper's final notation, the early-warning score is \(EW_{\Delta}=\bar{c}_1-\bar{s}_1\), so risk is flagged when \(EW_{\Delta}< -0.15\). The selected operating point corresponds to \(\theta_{\mathrm{ew}}=-0.15\) under this negative-risk convention and achieves F1 \(=0.327\). The full-conversation threshold \(\theta=-0.10\) is selected by the same procedure and achieves F1 \(=0.381\).}
  \label{fig:ew_threshold}
\end{figure}

The full-conversation threshold \(\theta=-0.10\) is selected by F1 maximization across thresholds from \(-0.05\) to \(-0.50\) on the full-conversation detection task. The early-warning threshold is selected by the same procedure on the midpoint early-warning task. The paper defines the early-warning score as \(EW_{\Delta}=\bar{c}_1-\bar{s}_1\), where lower values indicate that sustain-talk outweighs change-talk at midpoint. Figure~\ref{fig:ew_threshold} was generated using the equivalent earlier gap convention, \(\mathrm{gap}=\bar{s}_1-\bar{c}_1=-EW_{\Delta}\). Thus the plotted selected threshold \(-0.15\) should be read as the same illustrative operating point under the paper's negative-risk convention, \(EW_{\Delta}< -0.15\).

Because both thresholds are selected on the same dataset used for evaluation, all thresholded detection results should be interpreted as illustrative operating points rather than validated deployment thresholds.

\section{Connections to Existing Methods}
\label{app:connections}

The temporal signals used in the proof-of-concept connect to several established methodological traditions. These connections are not required for the main argument, but they clarify how richer SCOPE-MH operationalizations could be developed.

\paragraph{Survival analysis.}
First-sustain-dominance timing (\(T_{\mathrm{dom}}\)) is structurally analogous to time-to-event modeling in survival analysis: it tracks when a clinically meaningful state transition first occurs. Cox proportional hazards models could capture how therapist behavior patterns affect the hazard of sustain-talk onset, providing a principled alternative to the simple threshold-based heuristic used here.

\paragraph{Sequential pattern analysis.}
Maximum sustain streaks and temporal drift are instances of sequential pattern detection. Hidden Markov models could represent client motivational state as a latent variable evolving across turns, with sustain-talk dominance as an observable emission. Change-point detection algorithms would provide a more principled alternative to the midpoint split used in this proof-of-concept.

\paragraph{Motivational interviewing fidelity coding.}
The temporal signals are grounded in the Motivational Interviewing Treatment Integrity framework~\citep{moyers2016motivational}, which uses sequential coding of therapist and client behavior to assess session quality. SCOPE-MH operationalizes a subset of these principles, specifically client language change across the interaction, as computable temporal metrics. This connection supports the clinical interpretability of the signals: they are not arbitrary statistical constructs, but operationalizations of established quality criteria.


\newpage

\end{document}